\documentclass[preprint,12pt,authoryear]{article}
\usepackage{natbib}

\usepackage{color}

\usepackage{rotating}

\usepackage{listings}
\usepackage{graphics}
\usepackage{upgreek}
\usepackage{alltt}
\usepackage{amsmath}
\usepackage{amssymb}
\usepackage{amsthm}
\usepackage{calrsfs}
\usepackage{epsfig}
\usepackage[toc,page]{appendix}

\usepackage{subfigure}
\usepackage{amsbsy}
\usepackage{hhline}
\usepackage[para]{manyfoot}
\usepackage{multirow}
\usepackage{rotating}
\usepackage{array}
\usepackage{supertabular}
\usepackage{algorithmicx}
\usepackage[ruled]{algorithm}
\usepackage{algcompatible}
\usepackage{algpseudocode}
\algtext*{EndIf}

\usepackage{nicefrac}
\usepackage{xcolor}
\usepackage{soul}

\usepackage{amssymb}

\theoremstyle{plain}

\newcommand{\comment}[1]{}

\providecommand{\keywords}[1]{\textbf{\textit{Keywords ---}} #1}

\setcitestyle{aysep={},yysep={,}}

\DeclareMathAlphabet{\mathscr}{OT1}{pzc}{m}{it}

\title{An Algorithmic Inference Approach to Learn Copulas}

\author{
Bruno Apolloni \\
Department of Computer Science\\
University of Milano\\
Via Celoria 18, 20133, Milano, Italy \\
\texttt{apolloni@di.unimi.it} 
}

\begin{document}

\maketitle

  \begin{abstract}
  We introduce a new method for estimating the parameter $\alpha$ of the bivariate Clayton copulas within the framework of Algorithmic Inference. The method consists of a variant of  the standard bootstrapping procedure for inferring random parameters, which we expressly devise to bypass the two pitfalls of this specific instance:  the non independence of the Kendall statistics, customarily at the basis of this inference task, and the absence of  a sufficient statistic w.r.t. $\alpha$. The variant  is rooted on a numerical procedure in order to find the $\alpha$ estimate at a fixed point of an iterative routine. Although paired with the customary complexity of the program which computes them, numerical results show an outperforming accuracy of the estimates.

  \end{abstract}


   \section{Introduction}
Copulas are the atoms of the stochastic dependency between variables. For short, the Sklar Theorem~\citep{Sklar1973} states that we may split the joint cumulative distribution function (CDF) $F_{X_1,\ldots,X_n}$ of variables $X_1,\ldots,X_n$ into the composition of their marginal CDF $F_{X_i}, i=1,\ldots,n $ through their copula $C_{X_1,\ldots,X_n}$. Namely~\footnote{ By default, capital letters (such as $U$, $X$) will denote random variables and small letters
($u$, $x$) their corresponding realizations; 
 bold-faced characters will denote vectorial quantities. 
}:
\begin{equation}
\label{partition}
F_{X_1,\ldots,X_n}(x_1,\ldots,x_n)= C_{X_1,\ldots,X_n}\left(F_{X_1}(x_1),\ldots,F_{X_n}(x_n)\right)
\end{equation}
While the random variable $F_{X_i}(X_i)$ is a uniform  variable $U$ in $[0,1]$ for whatever continuous $F_{X_i}$ thanks to the Probability Integral Transform Theorem~\citep{Rohatgi1976} (with obvious extension to the discrete case),     
   $C_{X_1,\ldots,X_n}(F_{X_1}(X_1),$ $\ldots,F_{X_n}(X_n))$, hence $C_{X_1,\ldots,X_n}(U_1,\ldots,U_n),$ has a specific distribution law which characterizes the dependence between the variables.

For the former we rely on a statistical framework, called Algorithmic Inference (AI)~\citep{ApolloniEtAl2006}, allowing to infer parameters of the $X_i$ distribution law with a given confidence. 
In principle, we may infer parameters for whatever $F_{X_i}$, provided that we have statistics with specific properties  re questioned parameters -- which qualify them as \emph{well-behaving} statistics~\citep{ApolloniEtAl2010Joura} and are generally owned by the sufficient statistics. 

For the copulas things are more difficult, essentially because of two drawbacks:
\begin{enumerate}
\item The experimental data we refer to lead to a so called \emph{pseudo sample}. Namely,  in force of~(\ref{partition}) our basic statistics  to infer copula parameters are  the vectors $(\boldsymbol U_1,\ldots,\boldsymbol U_m)$,  where $\boldsymbol U_j=(U_{j1},\ldots,U_{jn})=(\widehat F_{\boldsymbol X_1}(x_{j1}),\ldots,\widehat F_{X_n}(x_{jn}))$ and $\widehat F_{X_i}(x_{ji})$ estimate of the marginals. Starting from $\boldsymbol X_1,\ldots,\boldsymbol X_m)$, we use the above   $\boldsymbol U$s  to compute the scalar statistics $\{T_1,\ldots,T_m\}$, where $t_i$ reckons the number of elements of the sample $\{(x_{11},\ldots,x_{1n}),\ldots,(x_{1m},\ldots,x_{mn})\}$ whose coordinates are all less than those of $(x_{i1},\ldots,x_{in})$, namely
\begin{equation}
\label{cum}
T_i=\#\left\{(X_{j1},\ldots,X_{jn}):X_{j1}<X_{i1},\ldots,X_{jn}<X_{in}\right\}/(m-1),\quad 1\leq i,j\leq m
\end{equation}
where $\#\{A\}$ denotes the cardinality of the set $A$. In essence,  the set $\{T_1,\ldots,T_m\}$ represents the lookup table of the empirical cumulative distribution function (ECDF) of the  $( F_{X_1}(X_1), \ldots,F_{X_n}(X_n))$ and of  $(X_1,\ldots,X_n)$, as well, so that each $T_i$ is not independent of the others. This is why we denote this set as a pseudo sample. \\
\item We generally do not have well-behaving statistics available. Namely, the distribution law of $C$ may assume a vast variety of shapes. In view of some symmetries we may expect in the phenomena we are studying, we generally focus on Archimedean copulas~\citep{McNeil2009}, defined as:
\begin{equation}
C_{X,Y}(u_1,\ldots,u_n)\equiv C_\phi(u_1,\ldots,u_n)=\phi^{-1}\left(\phi(u_1)+\ldots +\phi(u_n)\right)
\end{equation}
where $\phi$, the  copula \emph{generator}, is a decreasing convex function which is defined in $[0,1]$ such that $\phi(1)=0$. In none of these cases we have well-behaving statistics available (in the acceptation of \citep{ApolloniEtAl2006})  on which to base our inference of the copulas' distribution law.
\end{enumerate}

In this paper we propose statistical methods and numerical strategies to bypass these drawbacks in the special case of  bivariate copulas with known margins. In particular, we focus  on  the \emph{Clayton} subfamily~\citep{Genest1993}, which has both relatively elementary  generators and corresponding distribution of $T$ s. 
\emph{Per se}, this inference may represent a base level problem both as  to the parametrization, in contrast with non parametric instances ~\citep{BUCHER, Coolen-Maturi} and unknown margins instances~\citep{Genestt}, and as to the  dimensionality~\citep{HOFERT}. However, on the one hand, bivariate copulas are at the basis of many multidimensional copulas modelings, such as in~\citep{JSS}. On the other hand, the computation complexity functions of the solving algorithms maintain rather the same shapes~\citep{HOFERT}.

An early idea of our method was posted on the  blog~\citep{copu} some years ago in a followup to a NIPS poster session. Since then, the statistical framework has grown with extension to the multivariate random variables and parameters~\citep{ApolloniEtAl2010Joura, APOLLONI2018}. Thus, we think now is the proper time to explore this  idea in greater depth and to enhance its presentation with a  completer numerical analysis and methodological considerations. 
As a result, we obtain a bootstrap population of values of the parameter under estimation that are \emph{compatible} with the observed sample~\citep{abm}.  From this population we compute a point estimator  that is insensitive to the non-independence of the Kendall statistics and outperforming. We also compute confidence intervals that are not biased by asymptotic assumptions, whose coverages comply with the planned confidence levels.
As a notational remark, "AI" -- in our case the acronym for Algorithmic Inference -- coincides with the one for  Artificial Intelligence. While this paper provides  statistical tools with  obvious applications in the latter, to avoid confusion we declare that throughout the  text  the  AI acronym  refers exclusively to Algorithmic Inference.

The paper is organized as follows. In Section~\ref{secBoot} we introduce the bootstrap AI procedure to infer distribution parameters, describing in particular the special expedients we use to implement a proper variant which infers the unique parameter of Clayton Copulas. In Section~\ref{inplem} we describe the implementation of the corresponding numerical procedure. In Section~\ref{resul} we discuss the numerical results and the extensibility of the procedure to other families of copulas. Conclusions and forewords are the subject of   Section~\ref{concl}.

\section{Bootstrapping the parameter of the Clayton copulas} 
\label{secBoot}

\subsection{The basic bootstrapping instance:}
The standard bootstrapping procedure for estimating parameters within the AI framework is the following.

By modeling the questioned parameter of a random variable $X$ as a random variable $\Theta$ in turn (hence a random parameter for short), \emph{on the light} of an observed sample $\{x_1,\ldots,x_m\}$, we:
\begin {enumerate}

\item identify a sampling mechanism $\mathfrak M=(g_{\theta},Z)$ such that $X=g_{\theta}(Z)$ and $Z$, denoted as the seed,  is a completely known variable. For instance the unitary uniform random variable $U$ so that $g_{\theta}=F^{-1}_{X,\theta}$ for  $X$ continuous and analogous function for $X$ discrete~\citep{ApolloniEtAl2008a}). Accordingly, for $X$ negative exponential variable,we have  $F_X(x) =(1-e^{-\lambda x})I_{(0,\infty)}(x)$ and   $g_{\lambda}(u)=-\frac{Log(u)}{\lambda}$.

\item compute a meaningful statistics $S=h(X_1,\ldots,X_m)$ w.r.t.\ $\mathfrak M$. Meaningfulness is characterized by \emph{well behaving} properties in the AI framework. These properties  are  owned by  sufficient statistic $S=\sum_{i=1}^mX_i$ for the above negative exponential instance. More in general, well behavingness represents  a \emph{local} variant  of the sufficiency conditions, hence a relaxation of them, as detailed in~\citep{ApolloniEtAl2010Joura}.

\item derive the master equation by equating the leftmost and rightmost terms of the following chain:
$$ s=h(x_1,\ldots,x_m)=h\left(g_{\theta}(z_1,\ldots,z_m)\right)=\rho(\theta,z_1,\ldots,z_m)$$
 where $(z_1,\ldots,z_m)$ are the unknown seeds of $(x_1,\ldots,x_m)$. The chain reads $s=-\frac{\sum_{i=1}^m u_i}{\lambda}$, in the lead case.

\item draw samples  $(\widetilde z_1,\ldots,\widetilde z_m)$ from $Z$ and solve the master equation on $\theta$. Continuing the example,  we see that its solution is $\lambda=-\frac{\sum_{i=1}^m \widetilde u_i}{s}$ .
\end{enumerate}

Referring to~\citep{ApolloniEtAl2010Joura} for the theoretical proofs, by repeating the last step for a huge number $n$ of times we have data for building up the  $\Theta$ ECDF. From this distribution we may compute both central values such as mean or median to obtain a point estimator and confidence intervals of $\Theta$.

\subsection{Our bootstrapping instance}
Coming to copulas, we must further elaborate the algorithm.  For the  Clayton family we may exploit the tool suite in Table~\ref{suite}.

\begin{table}
\begin{center}
\begin{tabular}{|c|c|c|}
\hline
Item 		& $Role$ 			& Expression \\[0.1cm]
\hline
\cline{1-2}
\multirow{2}{*}{parameters} & parameter&$\alpha\in(0,\infty)$\\
& generator & $\phi_\alpha(u)=\frac{(u^\alpha-1)}{\alpha}$\\
\hline 
\cline{1-2}
\multirow{3}{*}{copula} & CDF&$C_\alpha(u_1,u_2)=(u_1^{-\alpha}+u_2^{-\alpha}-1)^{-1/\alpha}$\\
& PDF & $c_\alpha(u_1,u_2)=(\alpha+1)u_1^{-\alpha-1}u_2^{-\alpha-1}(u_1^{-\alpha}+u_2^{-\alpha}-1)^{-1/\alpha-2}$\\
& CDF$|u_1$ & $C_\alpha(u_2|u_1)=(v_1^{-\alpha}+u_2^{-\alpha}-1)^{-1/\alpha-1}v_1^{-\alpha-1}$\\
\hline 
\cline{1-2}
\multirow{2}{*}{Kendall fun.} & CDF&$K(t)=\frac{t(\alpha-t^\alpha+1)}\alpha I_{[0,1]}(t)$\\
& PDF &$k(t)=\frac{(\alpha+1)(t^\alpha-1)}\alpha I_{[0,1]}(t)$\\

\hline
\end{tabular}
\end{center}
\caption{The Clayton tool suite.}\label{suite}
\end{table}

The target is the parameter  $\alpha$. Of the copula distribution we will exploit the conditional cumulative distribution of $U_2$ given a value $u_1$ of $U_1$, namely $C_\alpha(u2|u1)$, so that from a pair of  seeds $\{v_1,v_2\}$ drawn from the independent $[0,1]$-uniform seeds $\{V_1,V_2\}$, we obtain a $\{U_1,U_2\}$ sample by inverting the equations
\begin{eqnarray}
v_1=u_1\\v_2=(v_1^{-\alpha}+u_2^{-\alpha}-1)^{-1/\alpha-1}v_1^{-\alpha-1}
\end{eqnarray}
hence
\begin{eqnarray}
\label{smec}
u_1=v_1 \\
u_2=\left(-v_1^{-\alpha }+\left(u_2v_1^{\alpha +1}\right)^{-\frac{\alpha }{\alpha +1}}+1\right)^{-1/\alpha }
\end{eqnarray}
We exploit  \emph{Kendall's function}  $K(t)$, i.e. the CDF of $T$, hence of $C(X_1,X_2)$, thanks to the  one-to-one correspondence with its Archimedean copula~\citep{Genest2011}).

\begin{figure}[!t]
\centering{
\includegraphics[width=.5\columnwidth]{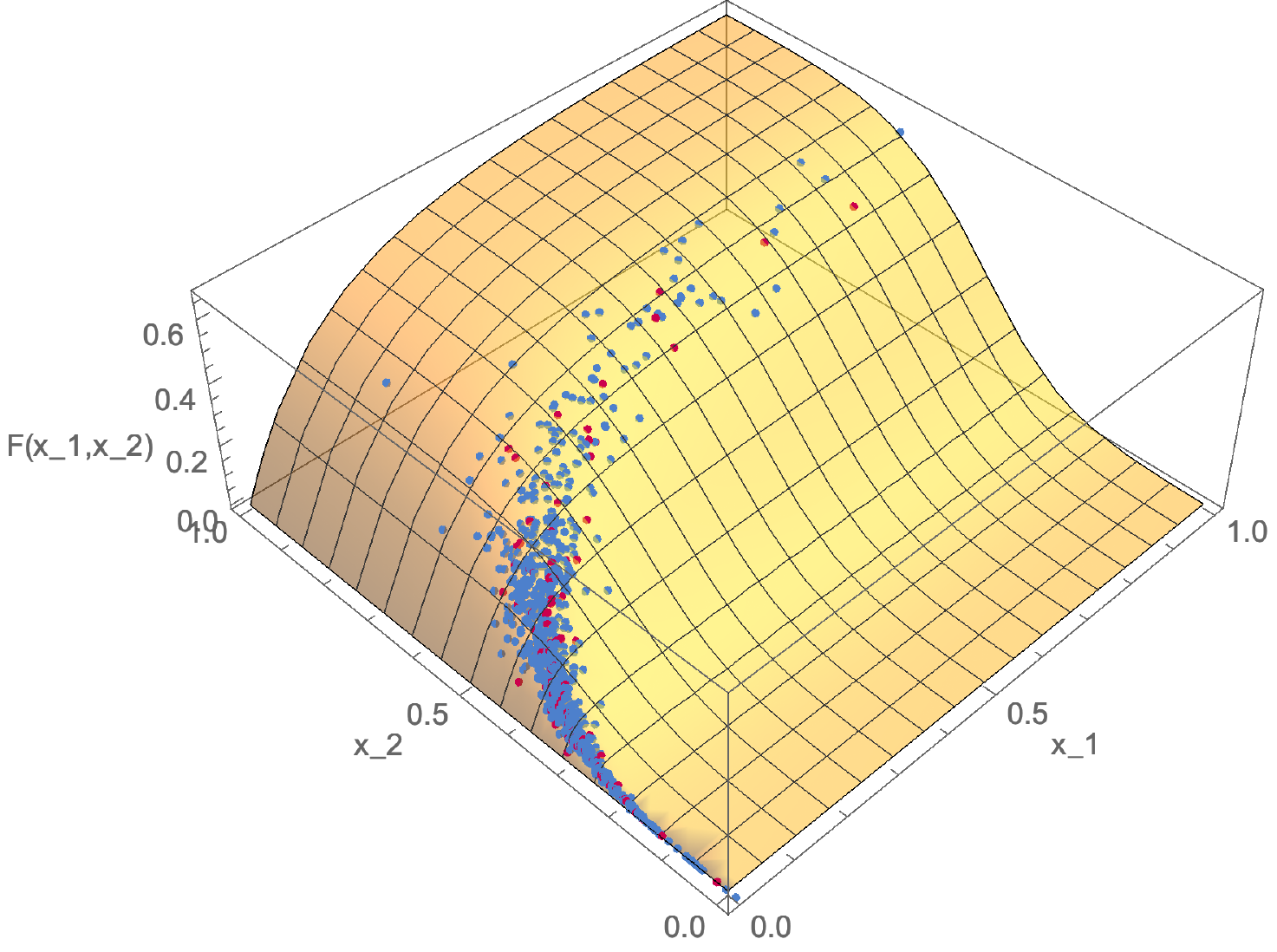}
\hspace{1 cm}
\includegraphics[width=.35\columnwidth]{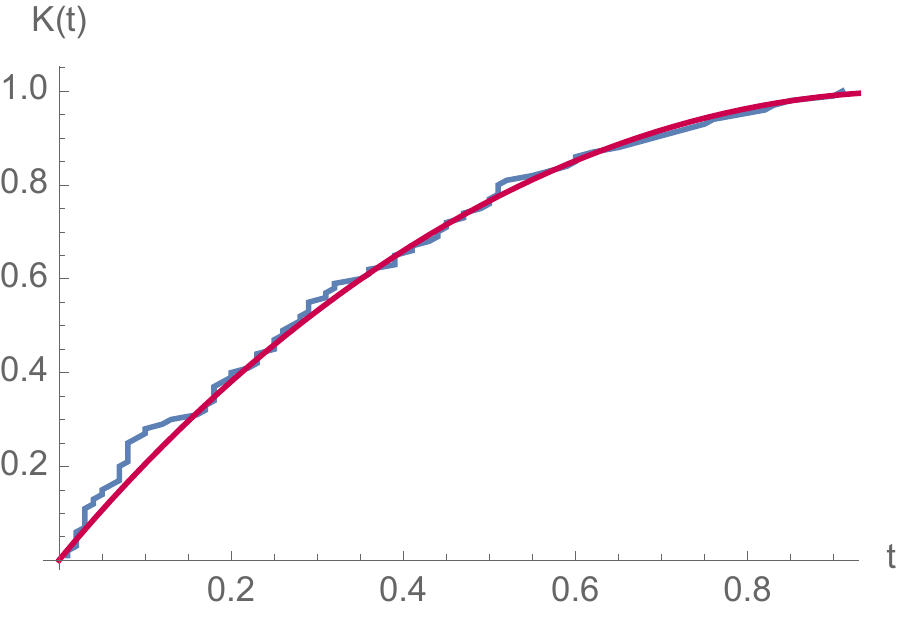}
}
\caption{A bivariate distribution PDF and its corresponding Kendall statistics empirical CDF \label{cox}}
\end{figure}

In Figure~\ref{cox}, by composing  a pair of variables $\{X_1,X_2\}$  --  respectively following   a Negative Exponential distribution with parameter $\lambda=44$ and a Gaussian distribution with parameters $\mu=0.5, \sigma=0.15$ --  through a Clayton copula with parameter $\alpha=0.8$, we obtain: a) the plot of the bivariate CDF via (~\ref{partition}) jointly with a sample of these variables  via~(\ref{smec}); b) the plot of  Kendall distribution CDF as detailed in Table~\ref{suite} and its empirical companion drawn  on the basis of a subsample of size $100$ of the above sample and their mapping in $T_i$ through (\ref{cum}).

 In order to identify the sampling mechanism (step 1 of our procedure) we consider its expression~\citep{Genest1993}:
\begin{equation}
\label{eqk}
u=\frac{t \left(1-t^{\alpha }+\alpha \right)}{\alpha } I_{[0,1]}(t)
\end {equation}

Our strategy is to infer $\alpha$ from a sample of $T$ derived from a sample of $\{X_1,X_2\}$. However,
notwithstanding the simplicity of the expression (\ref{eqk}) which depends uniquely on $\alpha$, no sufficient statistic exists for it. Hence to fulfill step 2, we decided to partition the CDF expressions so as to have two dummy distributions separately allowing a sufficient statistic whose expression is scarcely affected by the dependence between the sampled $T_i$s.
Namely, we consider the two \emph{dummy} CDFs:
\begin{equation}
\label{twineq}
\widetilde K_1(t)=t\frac{\alpha+1}{\alpha};\quad\quad \widetilde K_2(t)= \frac{t^{\alpha+1}}{\alpha}
\end{equation}
so that $s_1=\sum_{i=1}^m t_i$ and $s_2=\sum_{i=1}^m \log{t_i}$ are respectively the sufficient statistics for $\alpha$. The instantiation of the Integral Transform Theorem to identify $g^1_{\alpha}$ and $g^2_{\alpha}$ reads as follows.
\begin{equation}
U=W_1-(W_1-U)= T\frac{\alpha+1}{\alpha}-\frac{T^{\alpha+1}}{\alpha}
\end{equation}
where  we split the seed in two with the overall aim of having the original seed  $U$ facing (\ref{eqk}) in order to accomplish step 1.
From these equations we derive:
\begin{equation}
\label{sammec}
t =g^1_{\alpha}(w_1)= w_1 \frac{\alpha}{\alpha+1};\quad \quad t =g^2_{\alpha}(w_1-u)= \left(\alpha(w_1-u)\right)^\frac{1}{\alpha+1}
\end{equation}
Equating the two right members of (\ref{sammec}), 
for any sample $t_i$  we find the value of the second seed $w_1$ as a solution of the equation:
\begin{equation}
\label{eq8}
w_1\frac{\alpha}{\alpha+1}=(\alpha (w_1-u))^{1/(\alpha+1)}
\end{equation}
as a function of $u$. 

Now that the function $g_\alpha$ has been identified, at least in an implicit way, let us consider its seed. As previously mentioned, we are not working with independent $t_i$s. Thus we sample $U$ from the ECDF of $T$ CDF evaluated  on the sampled $t_i$s. This entails a circular procedure where, starting from a tentative $\widehat\alpha$ -- for instance its maximum likelihood estimate (MLE)~-- we evaluate the $w_{1i}$s so as to be able to implement steps 3 and 4 of our procedure by deriving $\widehat\alpha$ from the above sufficient statistics as:
\begin{equation}
\widehat\alpha_1=\frac{\sum_{i=1}^m t_i}{\sum_{i=1}^m w_{1i}-\sum_{i=1}^m t_i};\quad \widehat\alpha_2=\left\{\alpha:\sum_{i=1}^m \log(t_i)=\frac{\sum_{i=1}^m \log(w_{1i}-u_i) + m\log\alpha}{\alpha+1}\right\}
\end{equation}
Using  their mean $(\widehat\alpha_1+\widehat\alpha_2)/2$ as a new instance of $\alpha$ we may recompute $w_{1i}$s until convergence.

{\bf{Remark:}}
In another paper~\citep{APOLLONI2018} we acquainted a rather complementary inference problem on many parameters of a scalar variable distribution. In both cases we face a lack of independence on the involved statistics. The \emph{chainability} property,  invoked there as an antidote on the many parameters, here has a counterpart on the CDF split as in (\ref{twineq}).

\section{Implementing the bootstrap procedure}
\label{inplem}
The variant of the standard procedure we discussed in the previous section presents two distinguishing features as to the identification of the seeds and to their bootstrapping. Both entail computational problems that  we solved using standard tools available in a common mathematical package (Mathermatica 11.0 - Wolfram\citep{mmt}).
We are not interested in rescaling  twin CDFs in (\ref{twineq})  to reach exactly $1$ at their right extreme, since this does not affect the sufficiency of the  statistics $s_1$ and $s_2$, given the strictly monotone relationships between them and the parameter for common values of seeds and parameter. Rather, the crucial point of the procedure is the search for a fixed point for $\widehat\alpha$. This passes through a mean-field  process consisting of iterative solutions of~(\ref{eq8}) and a rough averaging of the two separate currently estimates of $\alpha$. 

To identify the  problems involved, we did a set of intermediate experiments. The experimental environment is represented by the pairs $\{\alpha,m\}$ in Table~\ref{tabexp}, tossed with samples of size $50$.

\begin{table}[ht]
\begin{center}
\begin{tabular}{|c|c|c|c|}
\hline
 $\alpha|m$  &20 & 30 & $100$ \\
\hline
0.8& 0.8 , 20 & 0.8 , 30 & 0.8 , 100 \\
1.7& 1.7 , 20&1.7 , 30 & 1.7 , 100 \\
3& 3 , 20 &3 , 30 &3 , 100 \\
5& 5 , 20 & 5 , 30 & 5 , 100 \\
\hline
\end{tabular}
\end{center}
\caption{The experimental plan. $\alpha \rightarrow$ target Clayton parameter; $m \rightarrow$ size of the sample processed by the estimators. }\label{tabexp}
\end{table}

\subsection{The parameter distribution}
\label{pd}
\begin{figure}[!t]
\begin{center}
\includegraphics[width=0.8\columnwidth]{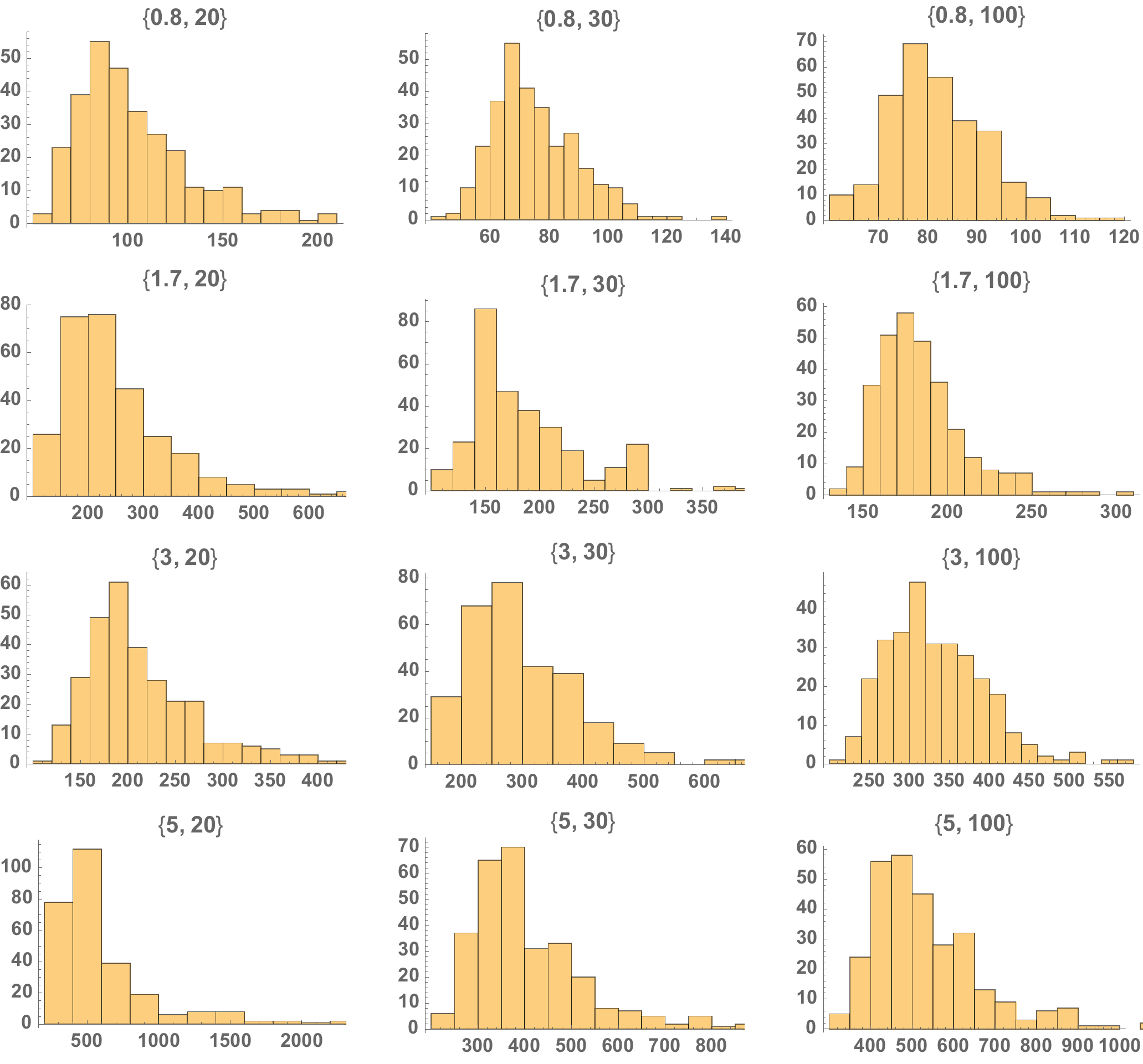}
\end{center}
\caption{An excerpt of the parameter distributions. The histograms refer to the value of the parameters times 100. \label{excerpt}}
\end{figure}

First, we use exactly  $\alpha$ and $m$ to generate both:
\begin{enumerate}\item the $50$ samples $(\{u_{1,i},u{_2,i}$ with $i\in\{1,\ldots,m\})$from $CDF|u_1$ in Table 1 
, $t_i$s from (\ref{cum}) and related statistics $(s_1=\sum t_i,s_2=\sum \log(t_i)$, and
\item for each sample a bootstrap population of $300$ replicas of  $m$ seeds $\{w_1,u-w_1\}$ from(\ref{eq8}) .
\end{enumerate}
From these \emph{exact} seeds we compute on each sample  the estimate $\hat\alpha=(\hat\alpha_1+\hat\alpha_2)/2$, so as to have 50 parameter populations of $300$ estimates on each cell of Table~\ref{tabexp}. While Figure~\ref{excerpt} reports a short excerpt of them,  Table~\ref{tabex} reports for each experimental cell the mean and standard deviation of the distribution central values. We contrast these values with the MLEs on the same $50$ samples on each cell, where MLE is directly computed by numerically maximizing the  product of the$t_i$  instantiations of Kendall PDF reported in Table~\ref{suite}. As customary in the  AI approach, central values are represented by the medians of   our estimators, contrasted with the single MLEs. We may see that our estimators generally outperform the  MLE companion.

\begin{table}[ht]
\begin{center}
\begin{tabular}{|c|c|cc|cc|cc|}
\hline
\multirow{2}{*}{ $\alpha$ } &statistic & $\alpha_{\mathrm{AI}}$&  $\alpha_{\mathrm{MLE}}$ & $\alpha_{\mathrm{AI}}$& $\alpha_{\mathrm{MLE}}$ & $\alpha_{\mathrm{AI}}$& $\alpha_{\mathrm{MLE}}$  \\
 && m=&20&m=&30&m=&100\\
\hline
\cline{1-2}
\multirow{2}{*}{0.8} &
mean&0.758288 & 1.9211&  0.756991 & 1.30998 &   0.794989 & 0.974008\\
&stdv&  0.0867599 & 1.39259 &  0.0426901 & 0.805682& 0.046596 & 0.266865 \\
\hline
\cline{1-2}
\multirow{2}{*}{1,7}
&mean& 1.57008 & 3.2841&1.62256 & 2.45001&1.68961 & 1.99565 \\
&stdv& 0.110117 & 2.53951& 0.109423 & 0.893306&  0.080659 & 0.47425\\
\hline
\cline{1-2}
\multirow{2}{*}{3}
 &mean& 2.72937 & 6.1436 & 2.84787 & 4.80562 & 2.89506 & 3.26851\\
 &stdv&0.247745 & 5.7625 & 0.222166 & 3.17289 &0.185544 & 0.808644\\
\hline
\cline{1-2}
\multirow{2}{*}{5}
 &mean&  4.42341 & 8.91136 &4.48985 & 9.16353& 4.7606 & 5.74891\\
 &stdv& 0.414849 & 7.69919 & 0.4082 & 8.90931&0.267965 & 1.78936\\
\hline
\end{tabular}
\end{center}
\caption{Comparison between a \emph{dummy} version of the proposed estimator $\alpha_{\mathrm{AI}}$ and maximum likelihood companion  $\alpha_{\mathrm{MLE}}$ computed on $50$ samples  for each instance of the experimental plan.  Cells: estimates' mean and standard deviation.}\label{tabex}
\end{table}

\subsection{The estimator distribution}
As mentioned before, we have no  \emph{true} $\alpha$ to draw the seeds $\{w_1,u-w_1\}$, hence we must replay it with an estimator within a circular procedure. We devote the first $300$ steps of the procedure to approach a fixed point $\alpha$ and another $300$ steps to collect a local distribution of $\hat\alpha$, whose median is used as the final estimator. Like the previous experiment, we compute this estimator on $50$ samples $(\{u_{1,i},u_{2,i}$ with $i\in\{1,\ldots,m\})$. On these vales we compute the same statistics as in the previous experiment, which we may see in Table~\ref{tabest}.

\begin{table}[ht]
\begin{center}
\begin{tabular}{|c|c|cc|cc|cc|}
\hline
\multirow{2}{*}{ $\alpha$ } &statistic & $\alpha_{\mathrm{AI}}$&  $\alpha_{\mathrm{MLE}}$ & $\alpha_{\mathrm{AI}}$& $\alpha_{\mathrm{MLE}}$ & $\alpha_{\mathrm{AI}}$& $\alpha_{\mathrm{MLE}}$  \\
 && m=&20&m=&30&m=&100\\
\hline

\cline{1-2}
\multirow{2}{*}{0.8} &
mean&  1.1177 & 1.93252 & 0.815224 & 1.30876&  0.792833 & 0.944016\\
&stdv&  0.694839 & 1.20238  &  0.43988 & 0.506298 & 0.196159 & 0.229342\\
\hline 
\cline{1-2}\multirow{2}{*}{1.7} &
 mean& 2.34081 & 2.81598 &  1.76855 & 2.32431& 1.71466 & 1.87771\\
&stdv& 1.87018 & 2.18893 & 0.956257 & 1.2762 &  0.358178 & 0.370696  \\
\hline 
\cline{1-2}
\multirow{2}{*}{3} &
 mean& 4.95876 & 7.15575  &  4.18669 & 5.26361&  3.06338 & 3.28108\\
&stdv&  2.41092 & 11.689 &  2.10042 & 6.02015  & 0.606473 & 0.737848\\
\hline 
\cline{1-2}
\multirow{2}{*}{5} &
 mean&  7.01755 & 7.54718&  6.30799 & 9.30124& 5.66863 & 5.84041\\
& stdv& 2.59534 & 10.8293& 1.37502 & 8.90931& 1.00474 & 1.7298\\
\hline
\end{tabular}
\end{center}
\caption{Comparison between the proposed estimator $\alpha_{\mathrm{AI}}$ and maximum likelihood companion  $\alpha_{\mathrm{MLE}}$. Same notation as in Table~\ref{tabex}.}\label{tabest}
\end{table}

 Since the initial value of  $\hat\alpha$ coincides  with the MLE on the same sample, this accounts for checking whether further computations improve the estimations or not. Though the actual estimators are less approximate than the dummy ones reported in the previous table, the edge of our procedure re MLE definitely remains as for both central values and dispersions. 
In the next section we will elaborate on this edge.

\section{Discussion}
\label{resul}

Willing to  explore the capability of our approach on the base problem of estimating the parameter $\alpha$ of the bivariate Clayton copulas, we leave to other papers (\cite{ApolloniEtAl2006,abm,ApolloniEtAl2010Joura})  the task of showing the comparative benefits and different semantics of AI approach re more assessed ones in the literature. Rather, in this section we constrain the discussion inside the AI  approach itself to appreciate the gain of further elaborations on data beyond the computation of the maximum likelihood, as for \emph{point estimators}. We also show our own implementation of the \emph{confidence intervals} for $\alpha$ and make considerations on the \emph{generality} of the proposed method.
\subsection{Point Estimates}
\begin{figure}[!t]
\begin{center}
\includegraphics[width=0.4\columnwidth]{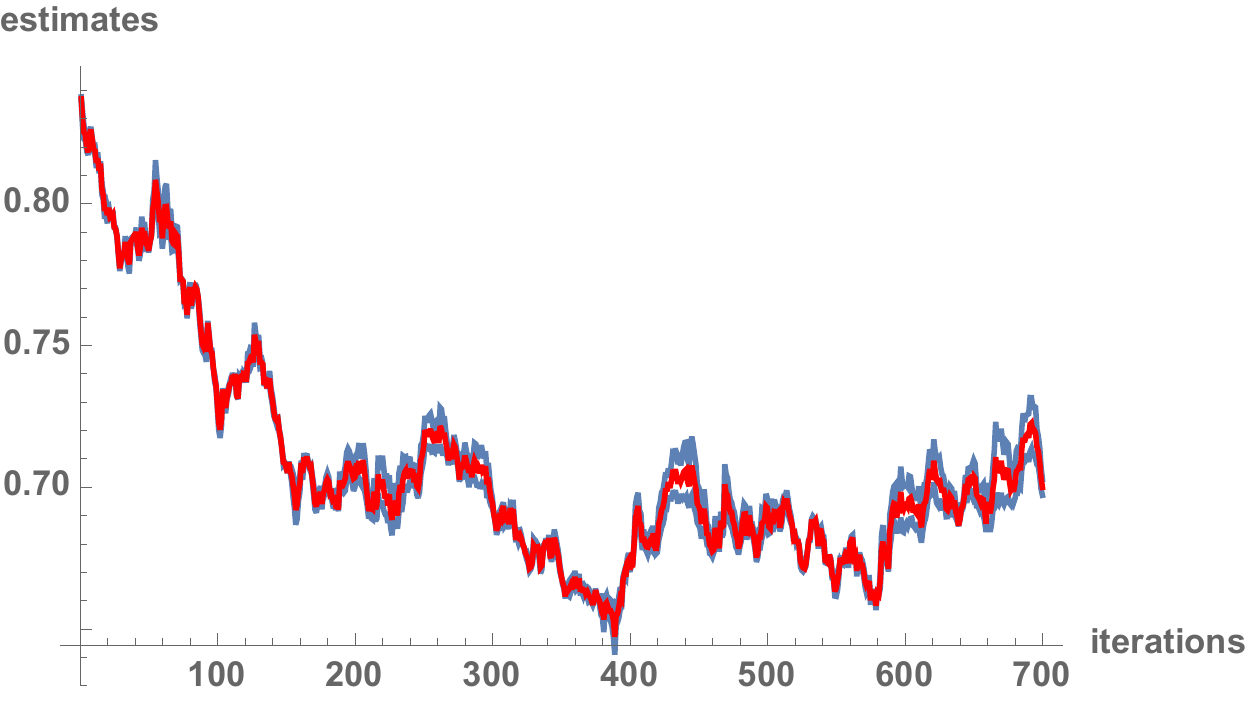}
\includegraphics[width=0.4\columnwidth]{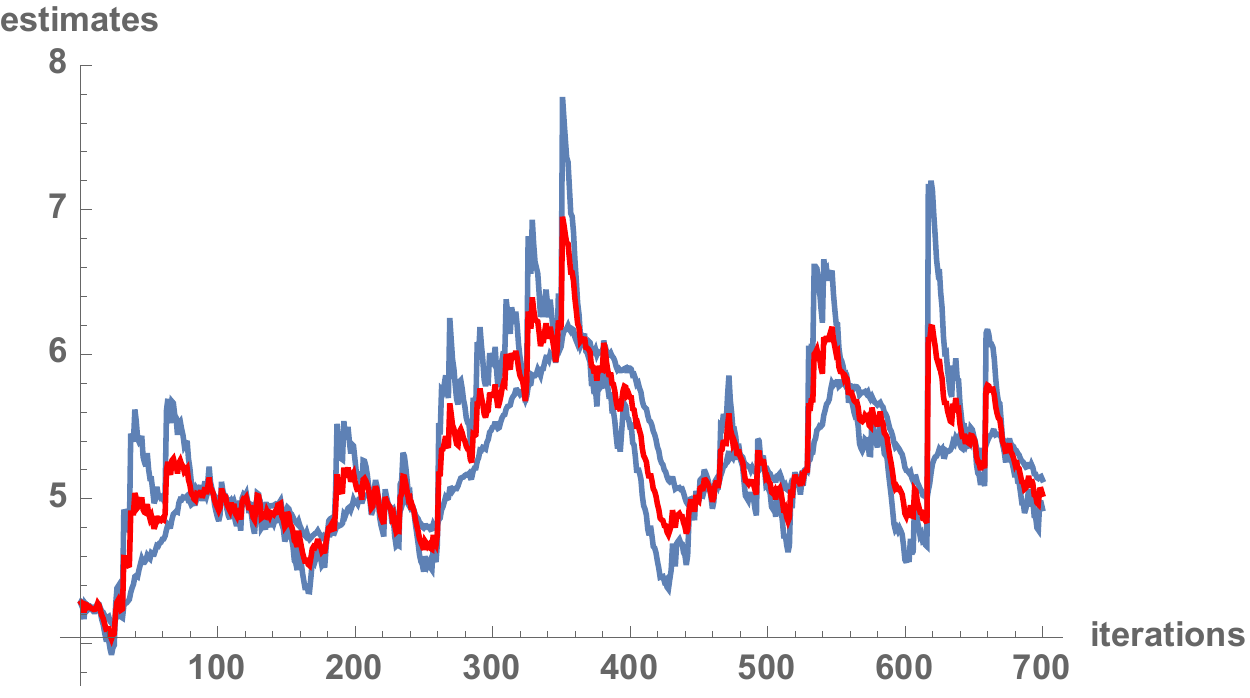}
\end{center}
\caption{Two mean-field  convergence  instances in the cells $(\alpha=0.8,m=20)$ and $(\alpha=5,m=100)$, respectively. Red path $\rightarrow \alpha_1$, blue path $\rightarrow \alpha_2$. \label{converg}}
\end{figure}

A  comparison between Tables~\ref{tabex} and ~\ref{tabest} highlights the drawback deriving from using an estimate of $\alpha$ in place of its true value during the generation of the pairs $\{w_1,u-w1\}$. This, in turn, derives from the bias  of  $\hat\alpha$. Actually, there is no great spread between $\alpha_1$ and $\alpha_2$ (see Figure~\ref{converg} for a typical example of joint trajectories along a cycle). 

\begin{figure}[!t]
\begin{center}
\includegraphics[width=0.4\columnwidth]{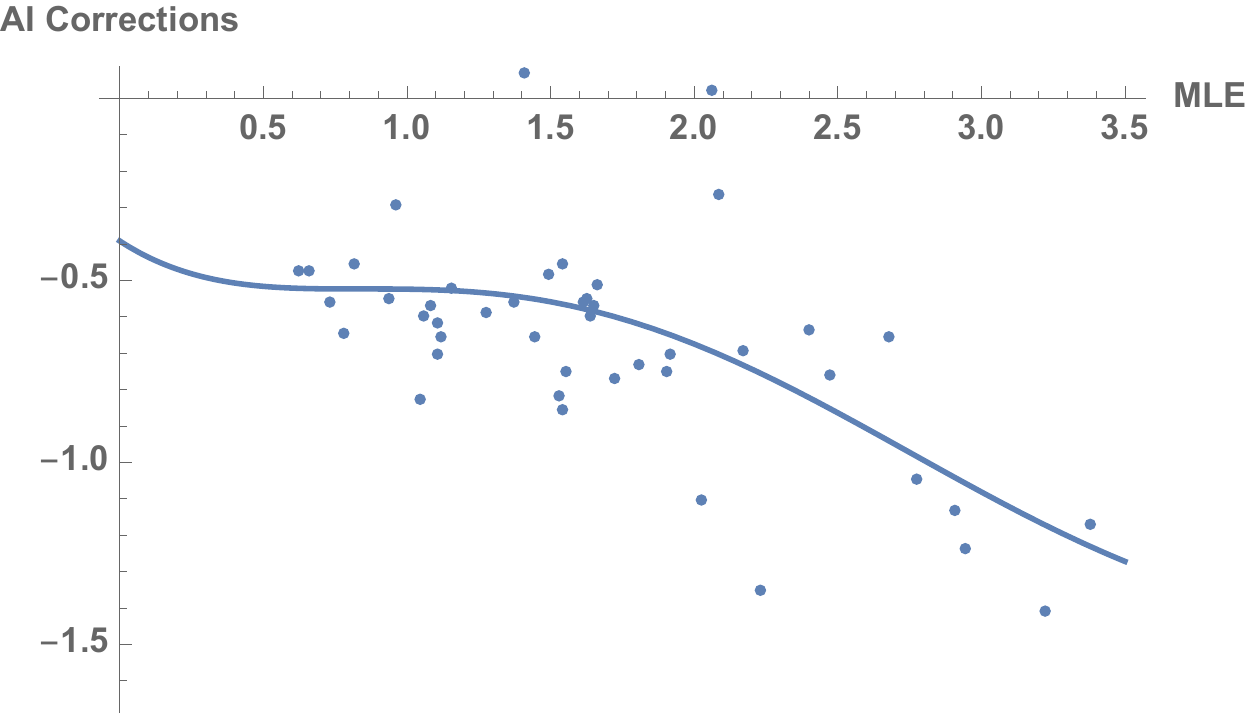}
\includegraphics[width=0.4\columnwidth]{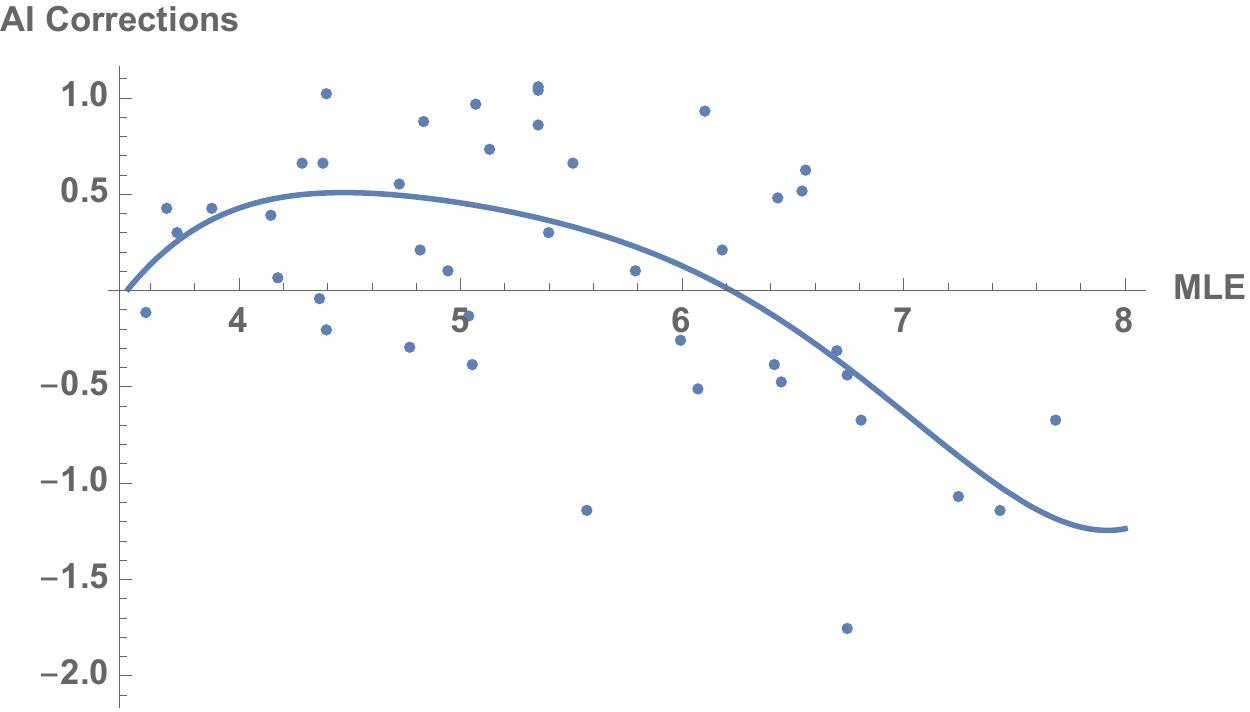}
\end{center}
\caption{The mean-field  corrections  in the cells $(\alpha=0.8,m=20)$ and $(\alpha=5,m=100)$, respectively. Continuous curve $\rightarrow$ fifth order fit. \label{correct}}
\end{figure}

Rather, the trajectories are banding, notwithstanding a low-coefficient-exponential smoothing adopted to constrain the oscillations, with the chance of being attracted by  local minima, especially for high vales of $m$.
Note that, MLEs too are spreading with respect to the target parameter, with the same trend re $m$. And, since these estimators are the starting point in the search for the $\hat\alpha$ fixed point,  MLE drifts generally induce analogous  $\hat\alpha$ drifts. Nonetheless, the mean-field trajectories generally get closer than MLE's to   the target. This is stemmed by the  individual corrections induced by the mean-field process on the original MLE, as shown in Figure~\ref{correct}. While for $\alpha=0.8$ we see a general bias toward lower values, with $\alpha=5$ the correction is more selective, by  inducing generally a positive shift in case of MLE underestimate and negative ones in the opposite case.

\begin{figure}[!t]
\begin{center}
\includegraphics[width=0.4\columnwidth]{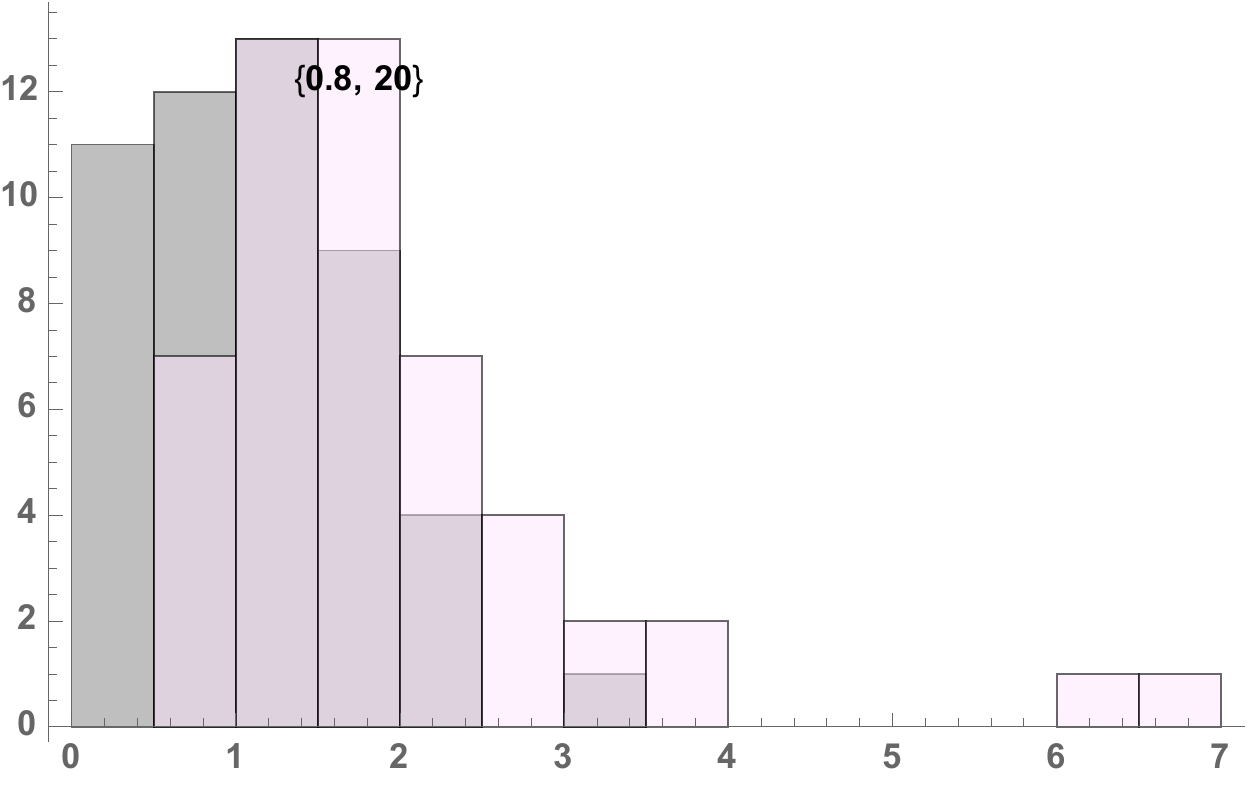}
\includegraphics[width=0.5\columnwidth]{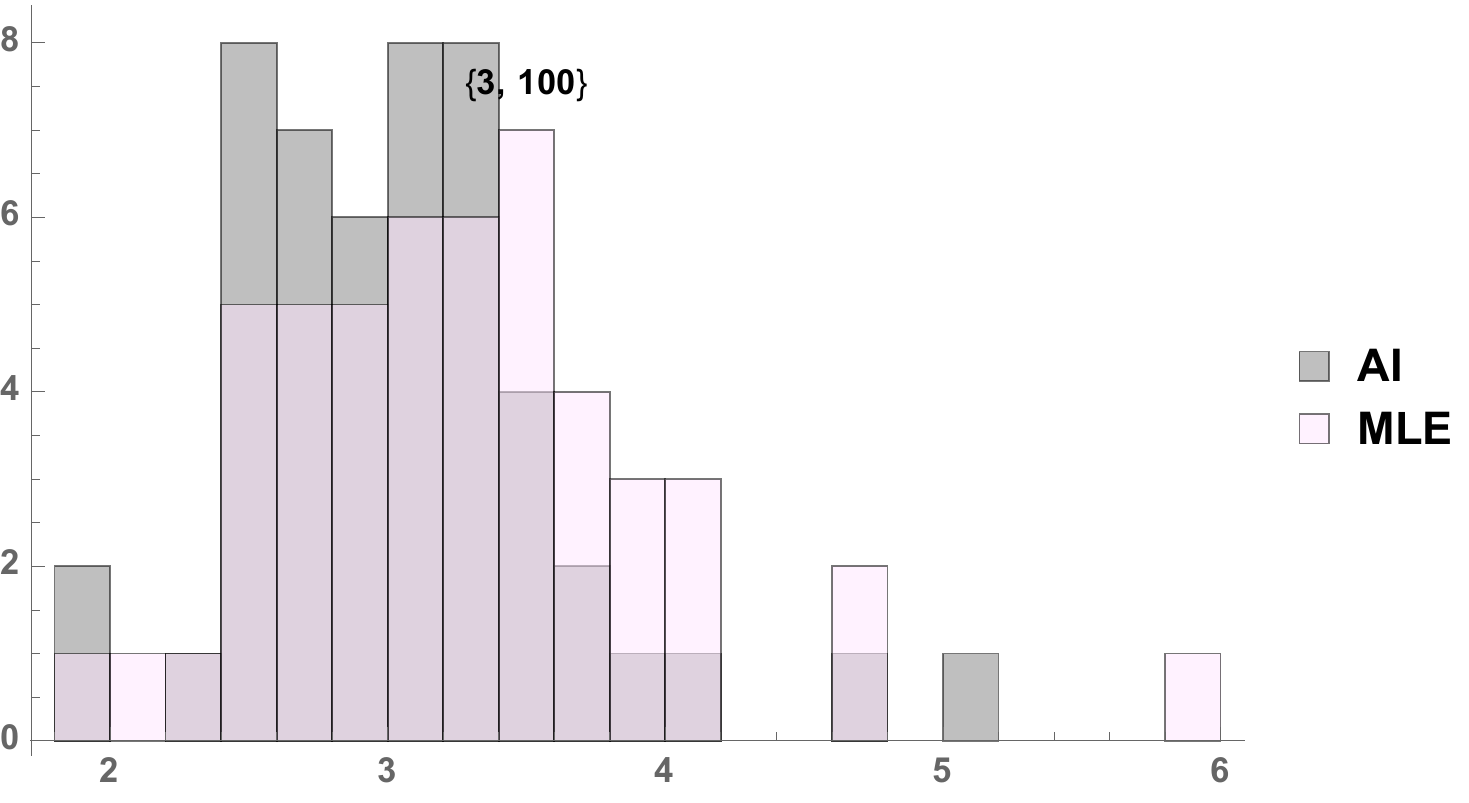}
\end{center}
\caption{Comparison of the histograms of the AI and MLE estimators in two typical instances. \label{istt}}
\end{figure}

As a result, we have a more favorable distribution of the estimates with our method. See Figure~\ref{istt} for a pair of instances. AI estimators are generally less biased and less dispersed. The biases are normally positive.

\subsection{Interval Estimates}
\begin{figure}[!t]
\begin{center}
\includegraphics[width=0.4\columnwidth]{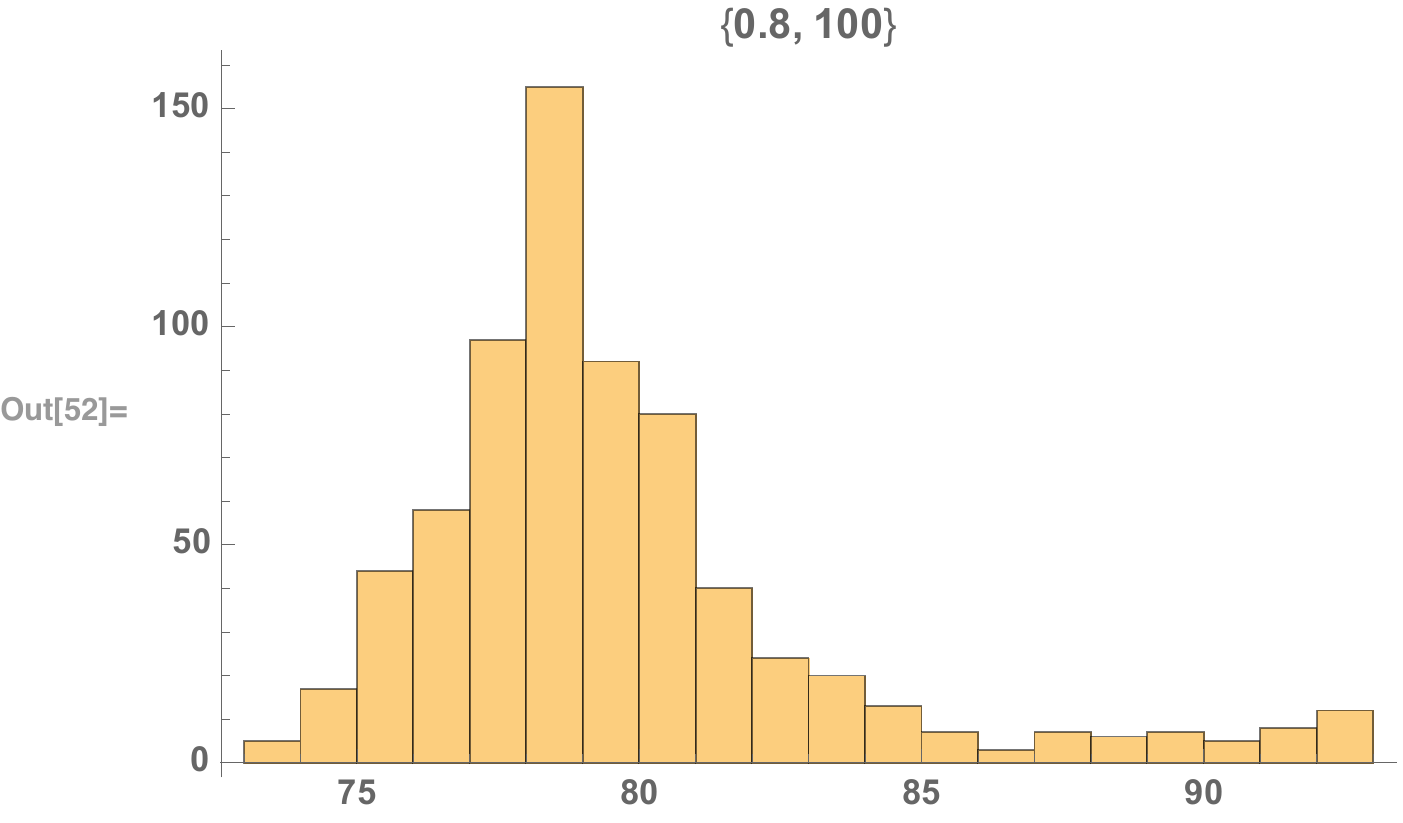}
\includegraphics[width=0.4\columnwidth]{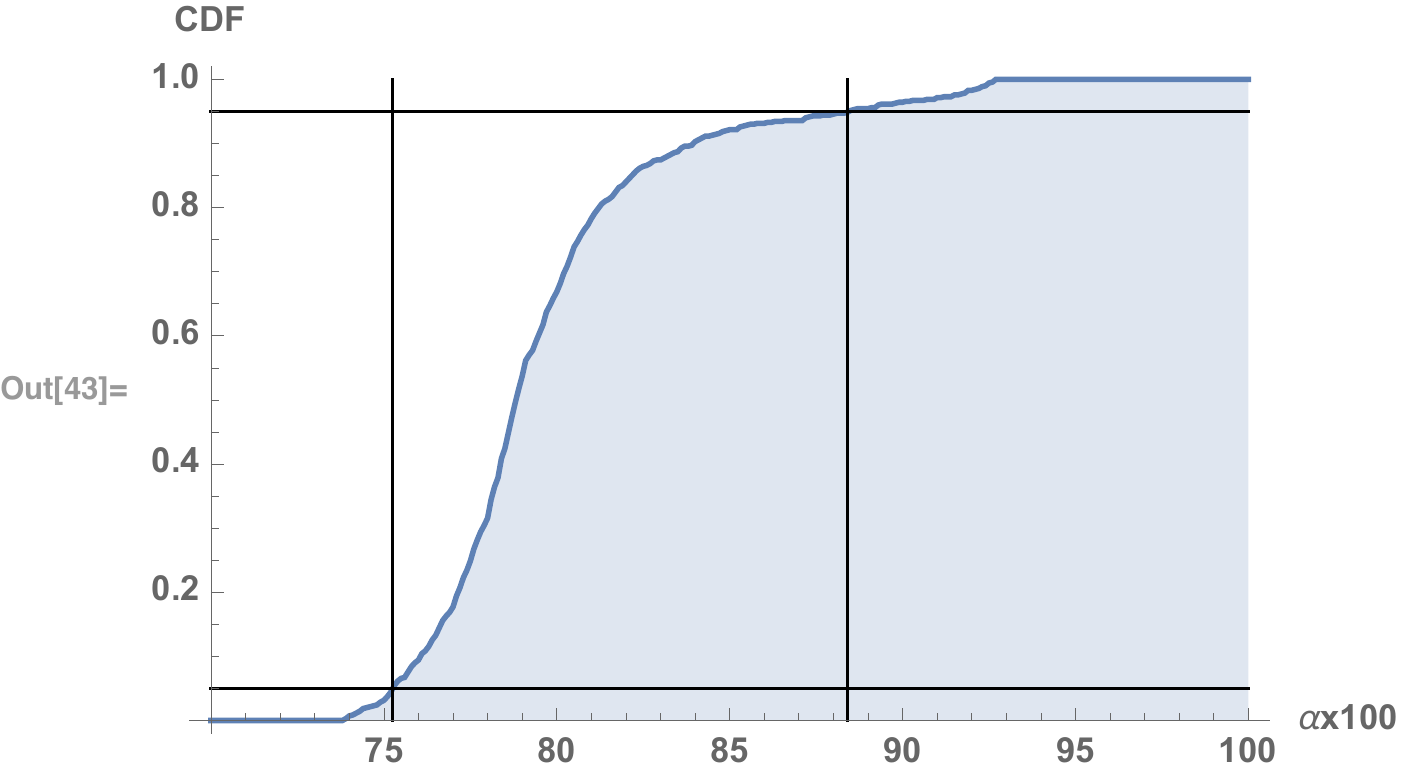}
\end{center}
\caption{From the histogram of the estimators\emph{ for a given sample} (left picture) to the quantiles identifying the confidence interval (right picture). To gain details, the estimate values are multiplied by $100$.  \label{istoveri}}
\end{figure}

Looking at the distribution of the $\hat\alpha$ replicas of our estimators we see non trivial empirical distributions (see Figure~\ref{istoveri}-left for instance) that can be used to compute confidence intervals for $\alpha$, yet avoiding to enforce an asymptotic  distribution approximation (normal, $\chi^2$, etc.)~\citep{CHEN,peng,HOFERT}. Namely, in the Algorithmic Inference framework we draw replicas of parameter estimates from replicas of their random seeds. By transferring the probability masses from the latter to the former, we obtain an empirical distribution of the random parameter that is \emph{compatible} with the observed sample~\citep{abm}, as in Figure~\ref{istoveri}-right. 

\begin{figure}[!t]
\begin{center}
\includegraphics[width=1\columnwidth]{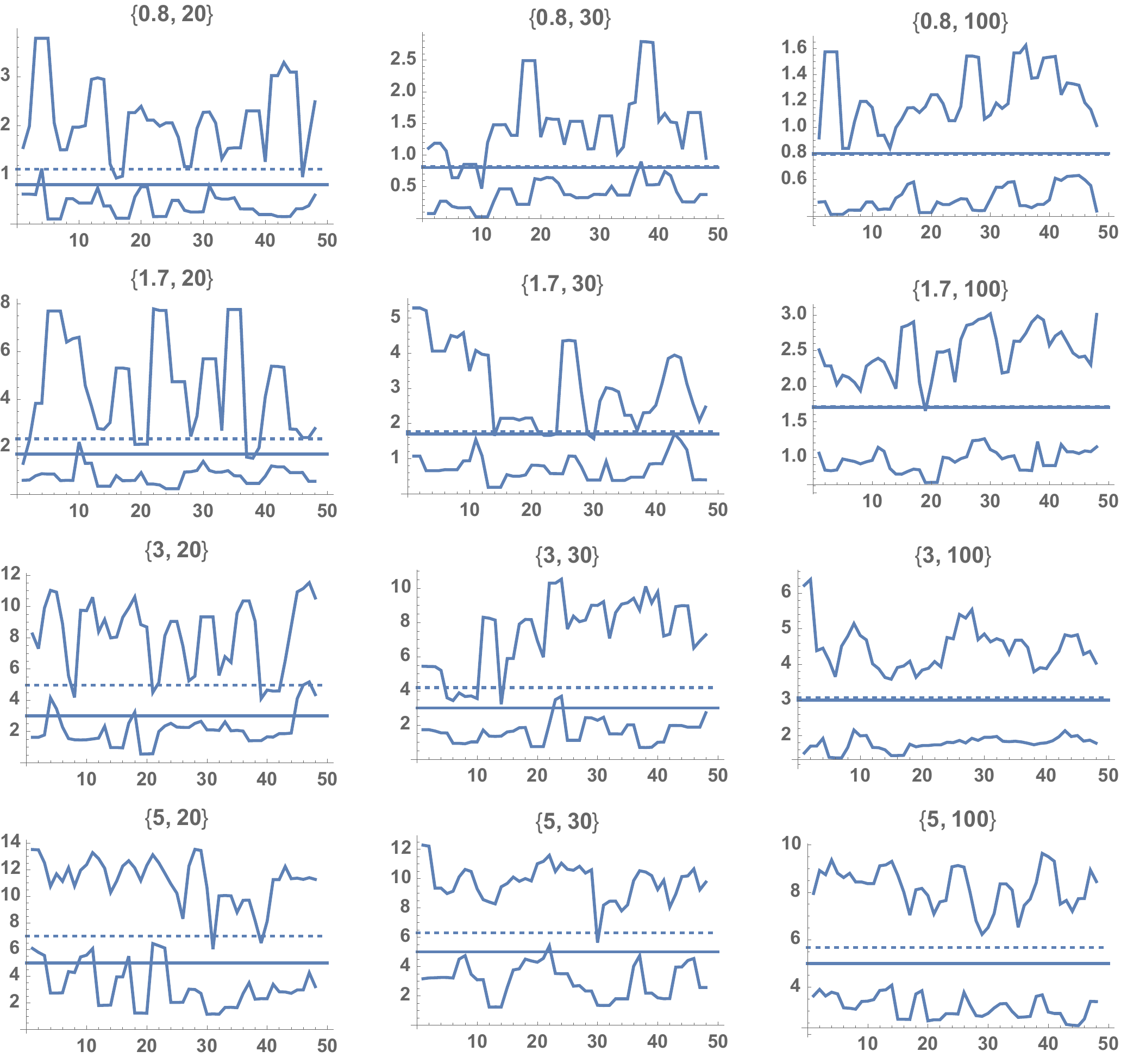}

\end{center}
\caption{Trends of the upper-bounds and lower-bounds of the $0.9$ confidence intervals with the samples in the various cells. Straight lines: continuous line  $\rightarrow$ original $\alpha$, dotted line estimated $\tilde\alpha$.   \label{intervalli}}
\end{figure}

Two expedients are necessary in order to get satisfactory intervals  by-passing two drawbacks in the statistic, respectively. 
\begin{enumerate}
\item Actually, assuming $\hat\alpha$ to be stabilized after a certain number of iterations, we could use the further tail of the mean-field process to draw $\alpha$ distribution. However, the continuous $\hat\alpha$ updating, and plus using the exponential smoothing, make the statistics highly correlated. Hence our strategy has been to exploit the tail to compute  a reliable estimate $\tilde\alpha$ as the median of the tail values,  to obtain a population of new estimates putting  $\tilde\alpha$ in (\ref{smec}) in analogy to what we did  in section~\ref{pd}. 
\item  Once  the parameter distribution  has been obtained, any confidence interval may be delimited by the proper quantiles of this distribution. However, in turn, the population of the new estimates suffers from the fact that the seeds $\{w_{1,i},u_i-w_{1,i}\}$ are based on $\tilde\alpha$ that is an approximation of the original value $\alpha$. This may entail hard shifts among the confidence intervals, which we reduce simply by  computing the quantiles on the merging of three consecutive empirical distributions (within the $50$ ones available on each cell). Figure~\ref{intervalli} shows the good coverage of the $48$ intervals that are obtained in this way for each cell, where the adopted quantiles are at levels $(0.05,0.95)$.
\end{enumerate}

\subsection{Procedure Extendability}
In conclusion, we may establish a  clear benefit deriving from our procedure. However, looking at the expressions of $T$ CDFs in Table~\ref{tabArchCop} we realize that only the Clayton and Gumbel copulas are easily separable as in (\ref{twineq}). Thus we expect to find some difficulties in extending  this procedure to other families of copulas.

\begin{table}[ht]
\begin{center}
\begin{tabular}{|c|c|c|}
\hline
Family 		& $\phi(v)$ 			& $K(v)$ \\[0.1cm]
\hline
Clayton		& $\frac{v^{-\alpha-1}}{\alpha}$	& $\frac{v(1-v^\alpha+\alpha)}{\alpha}$ \\[0.1cm]
Gumbel		& $(-\log v)^{\alpha+1}$		& $\frac{v(1-\log v)}{\alpha+1}$ \\[0.1cm]
Frank		& $\log\frac{1-\mathrm e^{-\alpha}}{1-\mathrm e^{-\alpha v}}$ & 
$v+\frac{1-\mathrm e^{-\alpha v}}{\alpha \mathrm e^{-\alpha v}}\log\frac{1-\mathrm e^{-\alpha}}{1- \mathrm e^{-\alpha v}}$ \\[0.1cm]
Joe		& $-\log(1-(1-v)^\alpha)$ & $v-\frac{\left(1-(1-v)^\alpha\right) (1-v)^{1-\alpha} \log \left(1-(1-v)^\alpha\right)}{\alpha}$ \\[0.1cm]
Ali-Mikhail-Haq & $\log \left(\frac{\alpha v-\alpha+1}{v}\right)$ & $v-\frac{(\alpha-1) \log \left(\frac{\alpha (v-1)+1}{v}\right)}{v (\alpha (v-1)+1)}$ \\[0.1cm]
\hline
\end{tabular}
\end{center}
\caption{Excerpt of Archimedean copula families with generator $\phi(v)$ and Kendall's function $K(v)$.}\label{tabArchCop}
\end{table}

\section{Conclusions}
\label{concl}

In this paper we use Algorithmic inference methods to solve the crucial problem of estimating the parameter $\alpha$ of the bivariate Clayton copulas. This task is  relevant in two respects: 
\begin{itemize}
\item from a functional perspective it is the gateway for dealing  with many dependency  estimates.
\item from a statistical perspective, the solution requires the engagement of non trivial algorithms to compute  the estimators.
\end{itemize}

The second aspect frequently occurs in  Computational Intelligence instances. The Algorithmic Inference framework faces it by explicitly exploiting the connections between the computational and probabilistic properties of statistics. In this way, we don't elude the computational burden of other inference methods such as MLE. Rather, we better finalize it to obtain functions that suitably transfer the statistical features  of a sample to the statistical properties of an unknown parameter. This passes through the identification of an ECDF of $\alpha$ that allows for suitable both point and by interval estimators.
Thanks to the numerical strategies discussed in this paper, 
the numerical results show that this approach provides comparative benefits that are tangible in both kinds of estimates.

\section*{References}

\bibliographystyle{chicago}
\bibliography{biblioprova}

\begin{thebibliography}{}

\bibitem[\protect\citeauthoryear{Apolloni and Bassis}{Apolloni and
  Bassis}{2011}]{ApolloniEtAl2010Joura}
Apolloni, B. and S.~Bassis (2011).
\newblock Confidence about possible explanations.
\newblock {\em {IEEE Transactions on Systems, Man, and Cybernetics, Part B:
  Cybernetics}\/}~{\em 41\/}(6), 1639--1653.

\bibitem[\protect\citeauthoryear{Apolloni and Bassis}{Apolloni and
  Bassis}{2018}]{APOLLONI2018}
Apolloni, B. and S.~Bassis (2018).
\newblock The randomness of the inferred parameters. a machine learning
  framework for computing confidence regions.
\newblock {\em Information Sciences\/}~{\em 453}, 239 -- 262.

\bibitem[\protect\citeauthoryear{Apolloni, Bassis, and Malchiodi}{Apolloni
  et~al.}{2009}]{abm}
Apolloni, B., S.~Bassis, and D.~Malchiodi (2009).
\newblock Compatible worlds.
\newblock {\em Nonlinear Analysis: Theory, Methods \& Applications\/}~{\em
  71\/}(12), e2883--e2901.

\bibitem[\protect\citeauthoryear{Apolloni, Bassis, Malchiodi, and
  Pedrycz}{Apolloni et~al.}{2008}]{ApolloniEtAl2008a}
Apolloni, B., S.~Bassis, D.~Malchiodi, and W.~Pedrycz (2008).
\newblock {\em The Puzzle of Granular Computing}, Volume 138 of {\em Studies in
  Computational Intelligence}.
\newblock Springer Verlag.

\bibitem[\protect\citeauthoryear{Apolloni, Malchiodi, and Gaito}{Apolloni
  et~al.}{2006}]{ApolloniEtAl2006}
Apolloni, B., D.~Malchiodi, and S.~Gaito (2006).
\newblock {\em Algorithmic Inference in Machine Learning, 2nd Edition}.
\newblock International Series on Advanced Intelligence, Vol. 5. Magill,
  Adelaide: Advanced Knowledge International.

\bibitem[\protect\citeauthoryear{blog copulas}{blog copulas}{2011}]{copu}
blog copulas (2011).
\newblock {\em Algorithmic Inference approach to learn copulas}.
\newblock http://www.probabilistic-numerics.org/apollonietal.pdf.

\bibitem[\protect\citeauthoryear{Brechmann and Schepsmeier}{Brechmann and
  Schepsmeier}{2013}]{JSS}
Brechmann, E. and U.~Schepsmeier (2013).
\newblock Modeling dependence with c- and d-vine copulas: The r package cdvine.
\newblock {\em Journal of Statistical Software, Articles\/}~{\em 52\/}(3),
  1--27.

\bibitem[\protect\citeauthoryear{Bücher and Volgushev}{Bücher and
  Volgushev}{2013}]{BUCHER}
Bücher, A. and S.~Volgushev (2013).
\newblock Empirical and sequential empirical copula processes under serial
  dependence.
\newblock {\em Journal of Multivariate Analysis\/}~{\em 119}, 61 -- 70.

\bibitem[\protect\citeauthoryear{Chen, Peng, and Zhao}{Chen
  et~al.}{2009}]{CHEN}
Chen, J., L.~Peng, and Y.~Zhao (2009).
\newblock Empirical likelihood based confidence intervals for copulas.
\newblock {\em Journal of Multivariate Analysis\/}~{\em 100\/}(1), 137 -- 151.

\bibitem[\protect\citeauthoryear{Coolen-Maturi, Coolen, and
  Muhammad}{Coolen-Maturi et~al.}{2016}]{Coolen-Maturi}
Coolen-Maturi, T., F.~P.~A. Coolen, and N.~Muhammad (2016).
\newblock Predictive inference for bivariate data: Combining nonparametric
  predictive inference for marginals with an estimated copula.
\newblock {\em Journal of Statistical Theory and Practice\/}~{\em 10\/}(3),
  515--538.

\bibitem[\protect\citeauthoryear{Genest, Neslehova, and Ziegel}{Genest
  et~al.}{2011}]{Genest2011}
Genest, C., J.~Neslehova, and J.~Ziegel (2011).
\newblock Inference in multivariate archimedean copula models.
\newblock {\em {TEST}\/}~{\em 20}, 223--256.

\bibitem[\protect\citeauthoryear{Genest and Rivest}{Genest and
  Rivest}{1993}]{Genest1993}
Genest, C. and L.~Rivest (1993).
\newblock Statistical inference procedures for bivariate archimedean copulas.
\newblock {\em Journal of the American Statistical Association\/}~{\em 88},
  1034--1043.

\bibitem[\protect\citeauthoryear{Genest and Segers}{Genest and
  Segers}{2009}]{Genestt}
Genest, C. and J.~Segers (2009).
\newblock Rank-based inference for bivariate extreme-value copulas.
\newblock {\em The Annals of Statistics\/}~{\em 37\/}(5B), 2990--3022.

\bibitem[\protect\citeauthoryear{Hofert, Mächler, and McNeil}{Hofert
  et~al.}{2012}]{HOFERT}
Hofert, M., M.~Mächler, and A.~McNeil (2012).
\newblock Likelihood inference for archimedean copulas in high dimensions under
  known margins.
\newblock {\em Journal of Multivariate Analysis\/}~{\em 110}, 133 -- 150.
\newblock Special Issue on Copula Modeling and Dependence.

\bibitem[\protect\citeauthoryear{Mathematica}{Mathematica}{2018}]{mmt}
Mathematica, w. (2018).
\newblock {\em Mathematica 11}.
\newblock http://www.wolfram.com.

\bibitem[\protect\citeauthoryear{McNeil and Neslehova}{McNeil and
  Neslehova}{2009}]{McNeil2009}
McNeil, A.~J. and J.~Neslehova (2009).
\newblock Multivariate archimedean copulas, d-monotone functions and l1-norm
  symmetric distributions.
\newblock {\em Annals of Statistics\/}~{\em 37\/}(5b), 3059--3097.

\bibitem[\protect\citeauthoryear{Peng and Ruodu}{Peng and Ruodu}{2014}]{peng}
Peng, L. and W.~Ruodu (2014).
\newblock Interval estimation for bivariate t-copulas via kendallÕs tau.
\newblock {\em Variance\/}~{\em 8\/}(1), 43--54.

\bibitem[\protect\citeauthoryear{Rohatgi}{Rohatgi}{1976}]{Rohatgi1976}
Rohatgi, V.~K. (1976).
\newblock {\em An Introduction to Probablity Theory and Mathematical
  Statistics}.
\newblock Wiley Series in Probability and Mathematical Statistics. New York:
  John Wiley \& Sons.

\bibitem[\protect\citeauthoryear{Sklar}{Sklar}{1973}]{Sklar1973}
Sklar, A. (1973).
\newblock Random variables, joint distribution functions, and copulas.
\newblock {\em Kybernetika\/}~{\em 9\/}(6), 449--460.

\end{thebibliography}

  \section*{Funding}
This research did not receive any specific grant from funding agencies in the public, commercial, or not-for-profit sectors.

  \end{document}